\newcommand{\bmvaOneDot}{. }

\documentclass[letterpaper, 10 pt, conference]{ieeeconf}  % Comment this line out if you need a4paper

\IEEEoverridecommandlockouts                              % This command is only needed if 
\title{\LARGE \bf
Synthetic Craquelure Generation for Unsupervised Painting Restoration
}

\author{Jana Cuch-Guillén$^{1,2}$, Antonio~Agudo$^{2}$, {\em Member, IEEE}, and Raül~Pérez-Gonzalo$^{2}$
\thanks{$^{1}$Universitat de Barcelona, Spain}
\thanks{$^{2}$Institut de Robòtica i Informàtica Industrial, CSIC-UPC, Spain}
}

\usepackage{amsmath,amsfonts}
\usepackage{algorithmic}
\usepackage{algorithm}
\usepackage{array}
\usepackage[font=normalsize,labelfont=sf,textfont=sf]{subcaption}
\usepackage{textcomp}
\usepackage{stfloats}
\usepackage{url}
\usepackage{verbatim}
\usepackage{graphicx}
\usepackage{amsmath}
\usepackage{cite}
\usepackage{booktabs}
\usepackage{etoolbox}
\usepackage{multirow}
\usepackage{caption}
\usepackage{float}
\usepackage{arydshln} % For dashed lines
\usepackage{hyperref} 
\hypersetup{
    colorlinks=true,
    linkcolor=blue,
    citecolor=blue,
    urlcolor=blue
}
\AtEndPreamble{
    \usepackage[capitalize]{cleveref}
    \crefname{section}{Sec.}{Secs.}
    \Crefname{section}{Section}{Sections}
\crefname{section}{Sec.}{Secs.}
\Crefname{section}{Sec.}{Secs.}

\crefname{subsection}{Sec.}{Secs.}
\Crefname{subsection}{Sec.}{Secs.}
    \Crefname{table}{Table}{Tables}
    \crefname{table}{Tab.}{Tabs.}
}

\def\etal{\emph{et al}\bmvaOneDot}

\begin{document}

\maketitle
\thispagestyle{empty}
\pagestyle{empty}

%%%%%%%%%%%%%%%%%%%%%%%%%%%%%%%%%%%%%%%%%%%%%%%%%%%%%%%%%%%%%%%%%%%%%%%%%%%%%%%%
\begin{abstract}
Cultural heritage preservation increasingly demands non-invasive digital methods for painting restoration, yet identifying and restoring fine craquelure patterns from complex brushstrokes remains challenging due to scarce pixel-level annotations. We propose a fully annotation-free framework driven by a domain-specific synthetic craquelure generator, which simulates realistic branching and tapered fissure geometry using Bézier trajectories. Our approach couples a classical morphological detector with a learning-based refinement module: a SegFormer backbone adapted via Low-Rank Adaptation (LoRA). Uniquely, we employ a detector-guided strategy, injecting the morphological map as an input spatial prior, while a masked hybrid loss and logit adjustment constrain the training to focus specifically on refining candidate crack regions. The refined masks subsequently guide an Anisotropic Diffusion inpainting stage to reconstruct missing content. Experimental results demonstrate that our pipeline significantly outperforms state-of-the-art photographic restoration models in zero-shot settings, while faithfully preserving the original paint brushwork.
%Our work can be found at:  \url{https://github.com/jana-cuch/PaintingRestorationIPCV}.
\end{abstract}

\section{INTRODUCTION}
Old paintings often develop cracks, or \textit{craquelure}, due to aging, environmental fluctuations, or mechanical stress~\cite{pizurica2015}. Craquelure not only signals deterioration but also distorts fine details and affects digital analysis~\cite{cornelis2013}, and while traditional restoration by conservators can be effective, it is irreversible and may further damage fragile artworks. These challenges motivate the development of non-invasive digital methods for crack detection and virtual restoration.

%Craquelure consists of extremely thin, irregular structures that vary widely across artworks, making them difficult to detect and separate from textured brushwork or aged varnish, and thus remains a persistent challenge for computational analysis and virtual restoration. Classical crack-detection methods can identify high-contrast fissures~\cite{abas2003,giakoumis2005}, but require important parameter tuning and struggle with textured surfaces, lightning and low-contrast cracks camouflaged in brushwork. The methods that rely on deep learning for the detection of cracks introduce improved robustness~\cite{sizyakin2020,sizyakin2022,nadisic2024}, but still, cracks' implicit nature of imbalance presence demands large annotated data, not present. Finally, modern transformed-based architectures require  with the thin, low-contrast structures that characterize cracks, as they usually tend to over-smooth them~\cite{vanvijle2025}. Taken together, these limitations highlight the need for a task-specific solution that can operate without manual annotations. 

Craquelure forms extremely thin, irregular, and low-contrast patterns that vary greatly across artworks, making them difficult to distinguish from textured brushstrokes and aged varnish layers. While classical crack detectors can recover high-contrast fissures, they require extensive parameter tuning and often fail under complex lighting or textured surfaces~\cite{abas2003,giakoumis2005}. Deep learning approaches improve robustness~\cite{sizyakin2020,sizyakin2022,nadisic2024}, yet their performance remains constrained by the scarcity of pixel-level annotations and the extreme class imbalance inherent to cracks. Moreover, recent studies indicate that modern transformer-based architectures may oversmooth thin structures and struggle to localize fine craquelure patterns~\cite{vanvijle2025}. These limitations highlight the need for a domain-specific strategy that can resolve fine cracks without relying on manually annotated training data.

%1.  classical issues: require substantial parameter tuning and struggle with textured paint surfaces, complex lighting and low-contrast cracks
%2. two practical problems persist: cracks are extremely thin and sparse, and precisely annotated training data are limited.
%3. but often struggle with thin, low-contrast structures typical of craquelures~\cite{vanvijle2025}.

\begin{figure}[!t]
\centering
\includegraphics[width=\linewidth]{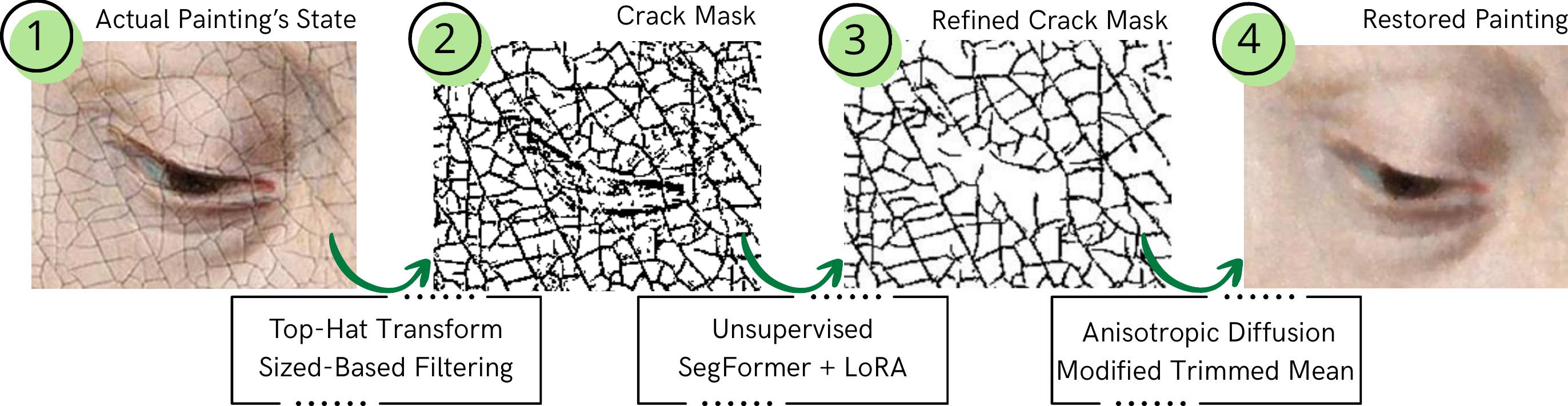}
\vspace{-0.5cm}
\caption{\textbf{Overview of the proposed unsupervised restoration algorithm.} Cracks are first identified using morphological filtering, then refined using a model fine-tuned on synthetic data, and finally restored via inpainting.}
\vspace{-0.45cm}
\label{fig:Proces scheme}
\end{figure}

To address these challenges, we propose a fully annotation-free framework for crack detection and virtual restoration, driven by a domain-specific synthetic craquelure model. First, an initial crack mask is extracted from the input painting using morphological top-hat filtering and size-based noise removal, which provides a coarse but over-inclusive set of candidate crack pixels. Because brushstrokes and textured pictorial elements are frequently misidentified at this stage, we introduce a learning-based refinement module that leverages a SegFormer~\cite{segformer} backbone with LoRA~\cite{lora} adaptation to suppress false positives and recover a clean crack map. This module is trained exclusively on our synthetic craquelure dataset. Finally, the refined mask guides a restoration stage in which crack pixels are filled using Modified Trimmed Mean (MTM) filtering and Anisotropic Diffusion (AD), enabling faithful reconstruction while preserving surrounding paint textures, as illustrated in Fig.~\ref{fig:Proces scheme}.

Thus, our work makes the following contributions:
\begin{itemize}
\item \textbf{Unsupervised crack detection and restoration.} We introduce the first framework that performs crack detection in paintings entirely without manual annotations. The system couples a classical morphological detector with a learning-based refinement module that uses detector-guided synthetic supervision to isolate genuine craquelure patterns from brushstrokes and texture.%, enabling reliable thin-structure localization under zero real labels.
%\item \textbf{Annotation-free crack detection.} We introduce a framework that performs crack detection entirely without manual annotations of real paintings. The system bridges the gap between classical and deep learning by using morphological priors to guide a synthetic-supervised refinement module.

\item \textbf{Synthetic craquelure dataset.} We generate artificial crack masks over clean paintings using Bézier trajectories, branched patterns, and tapered geometry. The resulting dataset produces a wide spectrum of craquelure morphologies, allowing models trained exclusively on synthetic data to generalize to real artworks.
%\item \textbf{Synthetic craquelure generator.} We generate artificial crack masks over clean paintings using Bézier trajectories with realistic branching and tapered geometry. This dataset captures a wide spectrum of craquelure morphologies, enabling the model to generalize to real artworks.

%We design through Bézier trajectories and tapered geometry inspired synthetic crack over non-damaged paintings. Our synthetic craquelure generator is explicitly designed to model a broad range of crack morphologies—including thin fractures, branched networks, fragmented fissures, and paint-layer separation patterns—so that the training set spans the diversity observed in real cracked paintings. This diversity is crucial for generalization: by exposing the model to many plausible crack types, we reduce overfitting to a single pattern and ensure robustness across artworks. This allows segmentation training entirely without real crack annotations.

\item \textbf{Integrated classical–learning restoration.} We present a unified restoration workflow that tightly integrates interpretable, classical detectors, a detector-conditioned segmentation refinement, and crack-aware inpainting.  This combination yields robust localization, reduced false positives, and visually coherent restorations while remaining computationally practical for conservation.

\end{itemize}

Experimental results show that we significantly outperform competitive annotation-free photo restoration models~\cite{wan2020,xu2023}. Qualitatively, we obtain sharper reconstructions that preserve original brushwork, confirming that our detector-guided strategy effectively isolates fine craquelure.

\newpage 
\section{RELATED WORK}

%The literature on computational and virtual restoration of paintings spans several directions that are relevant to the present study: classical image-processing crack detectors, deep-learning based crack detection and segmentation, multimodal imaging approaches, and virtual inpainting/restoration methods. Recent surveys provide useful overviews and highlight remaining challenges, most notably the detection and treatment of extremely thin, spatially sparse craquelure patterns in the context of limited annotated data~\cite{vanvijle2025,rathi2025}.

The literature on computational and virtual restoration of paintings spans several directions~\cite{vanvijle2025,rathi2025} that are relevant to the present study and highlight challenges in detecting and treating thin, sparse craquelure with limited annotated data.

\textbf{Classical crack detection and removal}.
Early approaches to crack detection in digitized paintings rely on handcrafted operators and classical image-processing pipelines. Morphological operators such as the top-hat transform and multi-scale filtering have been widely used to emphasize and extract thin dark cracks, followed by thresholding and chain-code or statistical descriptors to separate cracks from the background \cite{abas2003,giakoumis2005}. These methods can be effective for high-contrast cracks, but typically require substantial parameter tuning and struggle with textured paint surfaces, complex lighting and low-contrast cracks \cite{giakoumis2005,cornelis2013}.

\textbf{Deep learning for crack detection}.
Recent work leverages deep networks for crack detection and segmentation, eliminating the need for hand-crafted features \cite{wu2024}. Sizyakin \etal \cite{sizyakin2020,sizyakin2022} presented a learning-based detector tailored for multimodal painting imagery that improved performance over classical baselines. Nadisic \etal \cite{nadisic2024} proposed an active-learning framework (DAL4ART) which couples supervised training with active sampling and continuous retraining strategies. Other works have adapted popular classification backbones (e.g., VGG-16) to produce multi-class damage maps \cite{angheluta2020}. Despite these advances, two practical problems persist: cracks are extremely thin and sparse, and precisely annotated training data are limited.

%Recent work leverages Convolutional Neural Networks (CNNs) for crack detection and segmentation, eliminating the need for hand-crafted features \cite{wu2024}. CNNs have shown improved performance over classical methods, notably in multimodal painting imagery \cite{sizyakin2020,sizyakin2022} and through active-learning frameworks like DAL4ART \cite{nadisic2024}. Other approaches adapt popular classification backbones (e.g., VGG-16) to create multi-class damage maps \cite{angheluta2020}. However, practical challenges remain: cracks are extremely thin and sparse, and precisely annotated training data are limited.

\textbf{Multimodal imaging and domain-specific pipelines}.
Digital artwork analysis frequently leverages multimodal acquisitions (visible, IR, X-ray, etc.) to reveal subsurface features and boost detection \cite{cornelis2013,pizurica2015}. Fusion techniques range from sparse coding to learned architectures that explicitly exploit cross-modal correlations \cite{huang2020}. The downside is that multimodal data are not always available for every historical collection and can increase deployment complexity.

%Digital artwork analysis often uses multimodal acquisitions (visible, IR, X-ray) to reveal subsurface features and boost detection \cite{cornelis2013,pizurica2015}. Fusion techniques range from sparse coding to learned architectures \cite{huang2020}. The drawback is that multimodal data are not always available for all collections and can increase deployment complexity.

\textbf{Virtual restoration and inpainting}.
Beyond detection, many works address virtual crack removal and restoration.  Traditional inpainting based on non-local self-similarity and diffusion-based methods was adapted to artwork textures early on \cite{ruzic2010,ruzic2013}, but recent efforts have shifted towards generative models for better perceptual quality. For instance, the adaptive adversarial network was proposed to inpaint detected cracks, producing sharper results \cite{sizyakin2022}. Other systems combine detection and inpainting into staged pipelines to assist with non-invasive visualizations \cite{meeus2020}.

%Beyond detection, many methods address virtual crack removal. While early work adapted traditional inpainting (non-local self-similarity, diffusion) to artwork textures \cite{ruzic2010,ruzic2013}, recent efforts favor deep generative models for better perceptual quality. For instance, the adaptive adversarial network (aGAN) was proposed to inpaint detected cracks, producing sharper results \cite{sizyakin2022}. Other systems combine detection and inpainting into staged or end-to-end pipelines to aid conservators \cite{meeus2020}.

\textbf{Unsupervised strategies and cross-domain training}.
A recurring limitation in artwork restoration is the scarcity of annotated data. Prior work in other domains has shown that carefully constructed synthetic data or paired synthetic–real training strategies can provide strong priors for restoration tasks \cite{deeplearning_restoration_survey}. In the context of old-photo restoration, simulated degradations combined with photorealistic blending enables large-scale training without manual pixel-level annotation and improve transfer to real photos \cite{wan2020,xu2023}.

However, existing photographic restoration methods fail to capture the unique morphology of painting craquelure. Our work addresses this by introducing a domain-specific synthetic craquelure generator tailored to paint cracks, enabling the first fully unsupervised, synthetic-data-driven framework for crack restoration in cultural-heritage artworks.

%The scarcity of annotated data is a major limitation. Work in other domains, such as old-photo restoration, has shown that synthetic data and paired synthetic-real training can provide strong priors \cite{deeplearning_restoration_survey}. Simulated degradations paired with photorealistic blending allow large-scale training without manual pixel-level annotation and improve transfer to real photos \cite{wan2020,xu2023}.

\textbf{Image Segmentation}. Standard segmentation architectures (e.g., U-Net~\cite{unet}, DeepLabv3+~\cite{deeplabv3+}) are generally effective when cracks are relatively wide and sufficient annotated data are available. Transformer-based models~\cite{freqfusion,birefnet} including SegFormer~\cite{segformer}, as well as zero-shot universal segmenters~\cite{clipseg,diffseg} such as SAM2~\cite{sam2}, provide strong global context modeling~\cite{mask2former}, but often struggle with thin, low-contrast structures typical of craquelure. Specifically, recent studies suggest that their patch embeddings and multi-scale pooling tend to smooth or fragment fine spatial detail, underperforming on such structures~\cite{vanvijle2025}. These limitations motivated us to pursue a task-specific learning-based pipeline that leverages classical crack detection priors, reducing extreme class imbalance and preserving the high-frequency structures that transformer architectures typically attenuate.

\vspace{-0.05cm}
\section{UNSUPERVISED CRAQUELURE ART PAINTING RESTORATION} \label{sec:method}
\vspace{-0.05cm}

\subsection{Crack Detection} \label{sec:crack-detection}
Cracks in digitized artworks are generally visually characterized by low luminance and an elongated structure, allowing them to be modeled as local intensity minima. Acting on this premise,  we employ a preliminary crack detection step using morphological top-hat filtering, producing candidate crack masks that are subsequently refined and distinguished from real brushstrokes; as described in the next section.

\textbf{Top-hat transforms.} These transforms are particularly effective for isolating local contrast features by enhancing image components that stand out from their surroundings~\cite{top-hat}. In morphological terms, the \textit{black top-hat} transform highlights dark elements against a lighter background by computing the difference between the closing of an image and the original. Conversely, the \textit{white top-hat} transform emphasizes bright structures on dark backgrounds \cite{TopHatTransf}. Given that most cracks in art paintings appear darker than their surroundings, primary focus was placed on the black top-hat transform. The white variant was also tested to accommodate cases with inverse contrast. After top-hat filtering, the resulting image is thresholded by $180$ to produce a binary crack mask. %Thresholds were selected using Otsu’s method~\cite{otsu}, ensuring cracks are separated from background illumination variations.

%This implementation uses a flexible function to apply the top-hat operation with a chosen structuring element $B$. A $3 \times 3$ square and a disk of radius $2$ were tested; the disk better captured diagonal and thicker cracks, especially with more dilations. Applying $n$ successive dilations showed that small values (e.g., $n=1$) improve robustness, while larger values introduce artifacts (e.g., facial features), reducing segmentation precision. The optimal number of dilations depends on crack size: larger values highlight thicker cracks, while smaller ones preserve fine details, as shown in the upper row of Figure~\ref{fig:MTM, AD}.

%To improve the results, \textit{size-based filtering} was applied by removing connected components with fewer than 5 pixels. The lower this limit, the more noise the mask appears to have, giving worse results. This step helps to eliminate small noise elements that do not correspond to actual cracks. Components were labeled using $3 \times 3$ neighborhood connectivity, and their sizes were evaluated through a pixel count.

\begin{figure}[t]
\centering
\includegraphics[width=\linewidth]{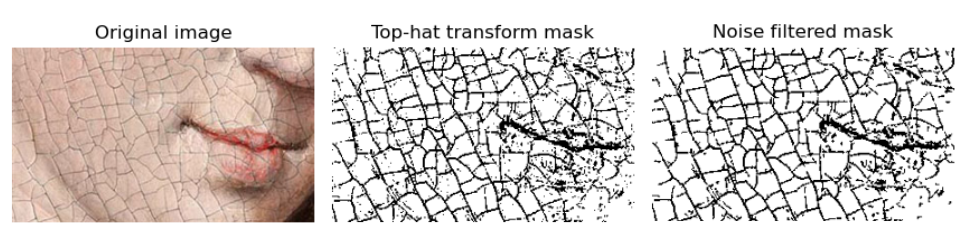}
\vspace{-0.5cm}
\caption{\textbf{Resulting top-hat mask} without noise filtering (center) and with noise filtering (right). }
\vspace{-0.5cm}
\label{fig:Crack detection}
\end{figure}

\textbf{Size-based filtering.} After binary segmentation, we filter out connected components smaller than 5 pixels to remove noise. This eliminates isolated artifacts that do not correspond to actual cracks, improving mask precision. We found that this small-object removal is crucial: setting the size threshold too low allows noise, while too high a threshold can omit real fine cracks. Components were labeled using $3 \times 3$ neighborhood connectivity and counted to enforce this filtering. Fig.~\ref{fig:Crack detection} shows the original image and two binarized results: one before and one after filtering. The filtered image provides a cleaner segmentation of cracks, as unnecessary features are largely removed.

%This implementation also incorporates interactive threshold and dilation sliders, allowing real-time tuning of the binarization and filtering thresholds. This proved useful for visually assessing the impact of different parameter values and adjusting them based on image content. Figures generated from the process are shown in Figure \ref{fig:Crack detection}, containing the original image and two binarized results: one before and one after filtering. The filtered image provides a cleaner segmentation of cracks, as unnecessary features are largely removed.

\vspace{-0.15cm}
\subsection{Learning-Based Brushstroke–Crack Segmentation}

%After the initial detection of cracks, the resulting crack mask usually includes many false positives, i.e., structures that visually resemble cracks but correspond instead to brushstrokes, canvas texture, or other pictorial features. To enhance the reliability of the restoration process, we introduce a secondary learning-based segmentation model whose objective is to refine this initial mask by suppressing false positives while preserving thin and low-contrast cracks.

%After the initial detection, the resulting crack mask usually includes many false positives, i.e., structures that visually resemble cracks but correspond instead to pictorial features. To enhance the restoration reliability, we introduce a learning-based segmentation model that refine the initial mask by suppressing false positives while preserving real cracks.

The initial crack mask often includes many false positives cracks that correspond to pictorial features. For this reason, we introduce a learning-based segmentation model to refine the mask, suppressing non-crack structures while preserving real cracks, thus enhancing restoration reliability.

\textbf{SegFormer with Parameter-Efficient Fine-Tuning.} Our refinement model is based on the SegFormer MiT-B0 architecture~\cite{segformer}, chosen for its transformer-based encoder and lightweight decode head. Instead of fully fine-tuning all parameters, we employ a parameter-efficient strategy using Low-Rank Adaptation (LoRA)~\cite{lora}. LoRA modules are injected into the attention and feed-forward projection layers, while only the LoRA parameters and the segmentation head remain trainable. This reduces the number of trainable parameters, enabling stable training on our synthetic data. Each attention matrix $W'\in \mathbb{R}^{d\times d}$ is decomposed as:
\begin{equation} \label{eq:lora}
W' = W + \Delta W, \qquad \Delta W = B A,
\end{equation}
where $A \in \mathbb{R}^{r \times d}$ and $B \in \mathbb{R}^{d \times r}$ are low-rank trainable matrices ($r \ll d$). During fine-tuning, $W$ is the pretrained weight, which is frozen, and only $A$ and $B$ are optimized. We adopt rank $r=8$ with LoRA
scaling factor $\alpha = 16$ and dropout of $0.1$.

\textbf{Detector-Guided Input Representation.}
To incorporate information from the initial crack detector, the model takes a four-channel input: the RGB image plus the detection map $m$ from \cref{sec:crack-detection}. The RGB channels are normalized using ImageNet statistics~\cite{imagenet}, while the detection channel is a binary mask marking candidate cracks. This extra channel guides the attention toward refining crack areas and discourages spurious predictions in homogeneous backgrounds.

%\textbf{Masked Hybrid Loss.} Since synthetic crack masks may emphasize certain regions more than others, we adopt a masked strategy that focuses training on areas where cracks are synthetically injected. Let $m \in \{0,1\}^{H \times W}$ denote the detector mask, where $H$ and $W$ are the height and width respectively, $y$ the ground truth cracks, and $p$ the predicted crack logit map. Rather than computing the loss over the entire image, we give higher weight to pixels where the detector is active:

\textbf{Masked Hybrid Loss.} Because synthetic crack masks may overemphasize certain regions, we use a masked loss to focus training where cracks are injected. Let $m$ be the detector mask, $y$ the ground truth, and $p$ the predicted crack logits. We define a weighting map:
\begin{equation}
w = m + \alpha (1 - m), \quad \alpha = 0.01,
\end{equation}
which prioritizes detector-activated areas while still allowing weak gradients elsewhere. With $\tilde{p}$ denoting the guided logit (see \cref{eq:guided-logit}), the final loss combines a weighted cross-entropy and a Dice loss~\cite{dice}:
\begin{equation} \label{eq:mask_loss}
\mathcal{L} = \mathcal{L}_{\text{CE}}(\tilde{p},y;w) + \lambda\, \mathcal{L}_{\text{Dice}}(p,y), \qquad \lambda = 2.
\end{equation} 

This hybrid loss encourages both local accuracy (CE) and global crack structure consistency (Dice), which is particularly important for extremely thin cracks where class imbalance is severe.

\textbf{Detector-Guided Logit Adjustment.} Beyond the masked loss, we also inject the detector mask into the model’s logits during training controlled by $\gamma\in \mathbb{R}$:
\begin{equation} \label{eq:guided-logit}
\tilde{p} = p + \gamma m, \qquad \gamma = 1.
\end{equation} 

This provides a soft bias toward predicting cracks where the detector indicates them, while allowing the model to correct false detections through supervision from the ground truth and the Dice loss~\cite{dice}.

\textbf{Training Setup.} Training is performed using AdamW~\cite{adamw} with a learning rate of $2\times 10^{-4}$ and a batch size of $8$. The training data is spatially augmented through horizontal/vertical flips, rotations, random crops, and scale–shift transforms. During validation, IoU is monitored for patience-based early stopping strategy~\cite{early}.

\vspace{-0.cm}
\subsection{Crack-Filling Methods}
After detecting all cracks and separating them from brushstrokes, the final step is to restore the damaged image using local information. We propose two complementary approaches, each applied independently to the RGB channels and restricted to pixels labelled as cracks, ensuring the rest of the image remains unchanged.

\vspace{0.15cm}
\textbf{Modified Trimmed Mean (MTM) Filter.}
This method offers a simple, fast, and effective way to inpaint crack pixels using local neighborhood statistics. Building on trimmed-mean strategies from the literature~\cite{MTMF}, we use a non-recursive outer-to-inner filling scheme that progressively restores crack pixels from the boundary inward. This ordering limits error propagation from heavily degraded areas and yields more consistent results.

For each crack pixel, the algorithm computes the mean of non-crack pixels in its 8-neighborhood, skipping pixels whose neighborhoods contain only cracks. After filling boundary pixels, the image is updated and the process repeats on previously skipped pixels until all cracks are inpainted, ensuring a coherent outer-to-inner restoration.

\textbf{Anisotropic Diffusion (AD).} AD is a Partial differential equation-based inpainting method that diffuses intensity along homogeneous directions while preserving edges and directional gradients. Its ability to follow image isophotes makes it suitable for filling wider cracks without blurring important structures.

A standard explicit finite-difference scheme following~\cite{AnisotropicDiff} is used. Let $I_{i,j}^t$ be the intensity at pixel $(i,j)$ at iteration $t$
\begin{equation}
    I_{i,j}^{t+1} = I_{i,j}^t + \lambda \hspace{-0.5cm} \sum_{k\in\{N,S,E,W\}} \hspace{-0.5cm} c_k D_k, \quad c_k = \frac{1}{1 + \left(\dfrac{|D_k|}{K}\right)^2},
    \label{eq:anisodiff} 
\end{equation}
where $D_k$ is the directional difference ($D_N = I_{i-1,j}^t - I_{i,j}^t$) and $c_k$ is the conductivity (edge-stopping). We set the time step to $\lambda = 0.25$ and gradient-sensitivity to $K = 127$.

The diffusion process is restricted solely to the pixels covered by the crack mask. Pixels outside the crack region remain fixed and act as Dirichlet boundary conditions to guide the inpainting process. We iterate Eq.~\eqref{eq:anisodiff} for a fixed number of steps (20 steps).

\begin{figure}[t!]
\centering
\includegraphics[width=\linewidth]{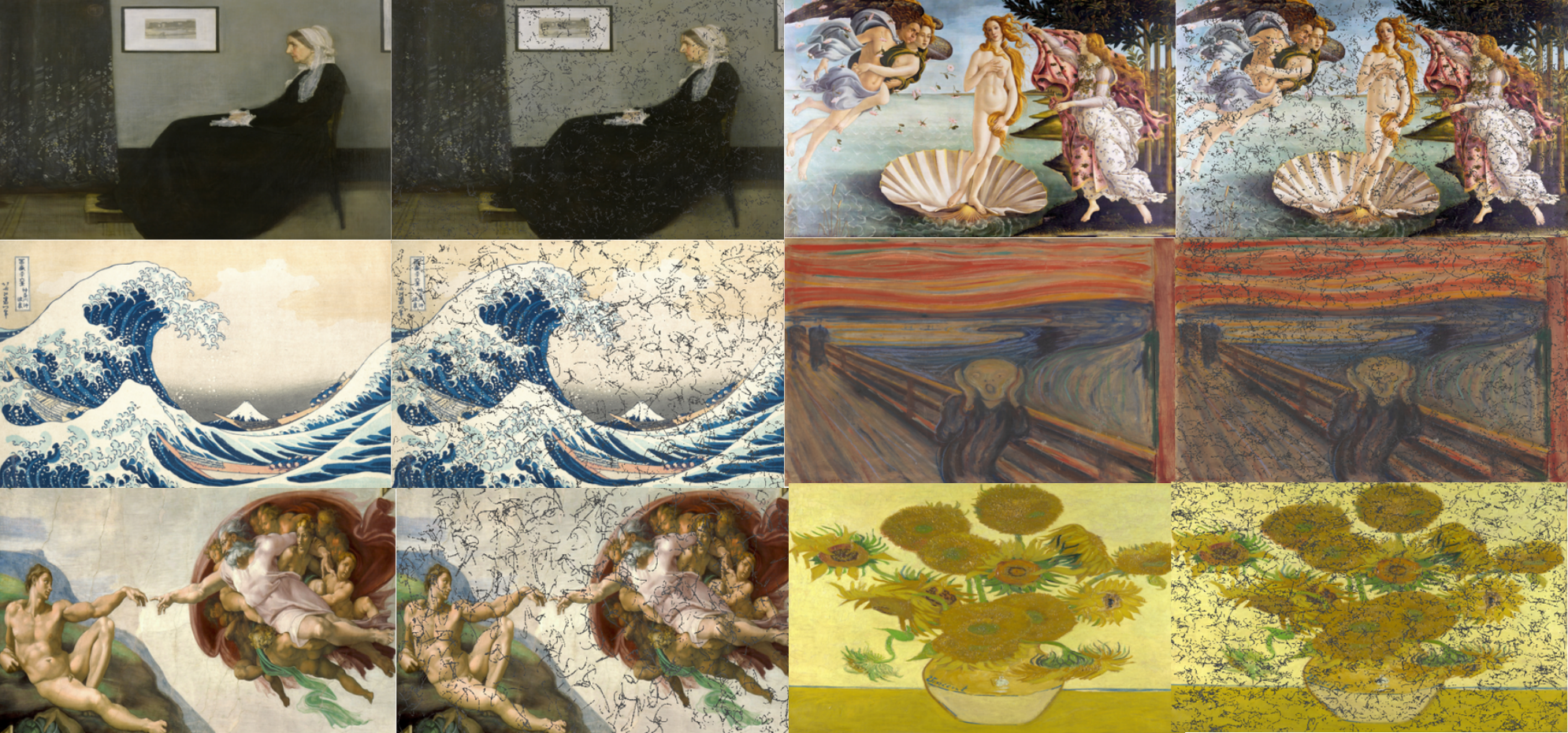}
\vspace{-0.55cm}
\caption{\textbf{Example of synthetic generated data}: pairs of original paintings (left) and resulting damaged images (right).}
\vspace{-0.4cm}
\label{fig:data}
\end{figure}

\begin{figure}[t!]
\centering
\includegraphics[width=\linewidth]{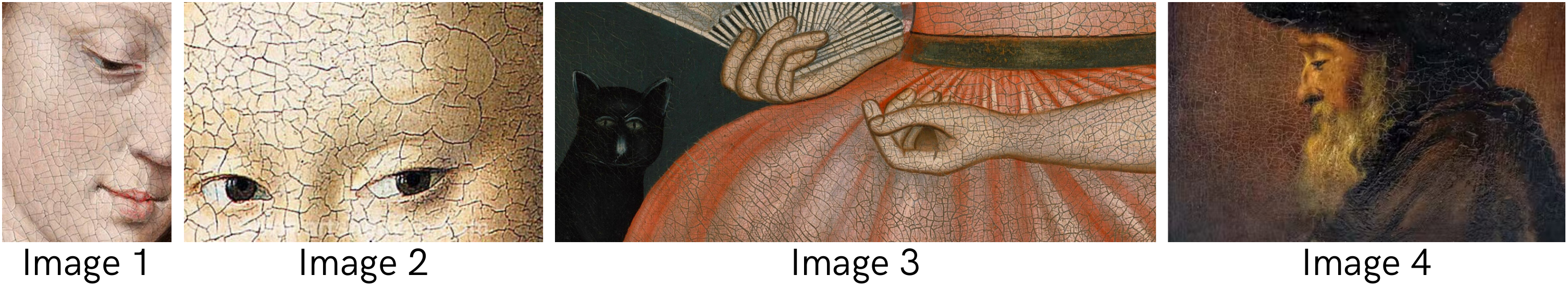}
\vspace{-0.6cm}
\caption{\textbf{Paintings with real cracks} constituting our test set.}
\vspace{-0.5cm}
\label{fig:test-data}
\end{figure}

\section{SYNTHETIC CRAQUELURE DATA} \label{sec:dataset}

To reduce reliance on costly manual annotation, we generate a synthetic train dataset of cracked paintings as aligned triplets \((I, M, \tilde{I})\): a clean WikiArt~\cite{wikiart} image \(I\), a binary crack mask \(M\), and the corresponding damaged image \(\tilde{I}\). Sample generations are shown in \cref{fig:data}. To evaluate real deterioration, four real-world paintings (\cref{fig:test-data}) were manually annotated and digitally restored, forming our final test set.

%This enables supervised training while keeping the entire pipeline unsupervised with respect to real cracks.

%\textbf{Crack Geometry Model}. 
We model crack trajectories with cubic Bézier curves~\cite{agudoICIP21}, defined by endpoints $p_0, p_3$ and internal control points $p_1, p_2$ perturbed by Gaussian noise to introduce curvature:
\begin{equation}
B(t)=(1-t)^3p_0 + 3(1-t)^2 t\,p_1 + 3(1-t) t^2 p_2 + t^3 p_3,
\end{equation}
with $t\in[0,1]$. The curve is uniformly sampled into 80 to 180 points, depending on its length. Endpoints $p_0,p_3$ are chosen uniformly in the image domain; $p_1,p_2 \sim \mathcal{N}(0, \sigma_p^2=8^2\text{ px})$.

%\subsection{Thickness Modulation and Rasterization}
Real cracks exhibit tapered geometry, with thinner extremities and thicker mid-sections. To emulate this behavior, at each sample \(B(t)\) we draw a filled disk of radius:
\begin{equation}
r(t)\sim\mathcal{N}\!\big(\alpha(1-|t-0.5|),\ \sigma_r^2\big),
\end{equation}
which yields a tapered profile (with \(\alpha=2.0\) px, \(\sigma_r=0.5\) px). Repeating this process for 80--150 curves per image produces dense and varied crack networks. To increase variability, with probability \(p_{\text{br}}\in[0.3,0.5]\), a branch is spawned by rotating and scaling the local direction vector.

%\subsection{Mask Post-processing}
The raw rasterized crack mask typically exhibits overly sharp boundaries. To improve realism, two refinement steps are applied: (i) \(2\times 2\) morphological erosion to refine thickness, and (ii) gaussian blurring with a $5\times5$ kernel ($\sigma=2$) to soften edges and mimic pigment bleeding. 

The blurred mask is thresholded at 50 to obtain a binary crack mask $M$. A damaged image $\tilde{I}$ is generated by replacing pixels in $M$ with a crack-specific gray value, leaving other pixels unchanged. This simulates paint loss and dark fissures in aged varnished paintings. All images are then resized (e.g., $598\times375$) for downstream segmentation.

\vspace{-0.15cm}
\section{RESULTS}
\vspace{-0.15cm}

\subsection{Crack Detection}

\begin{figure}[t!]
\centering
\includegraphics[width=0.85\linewidth]{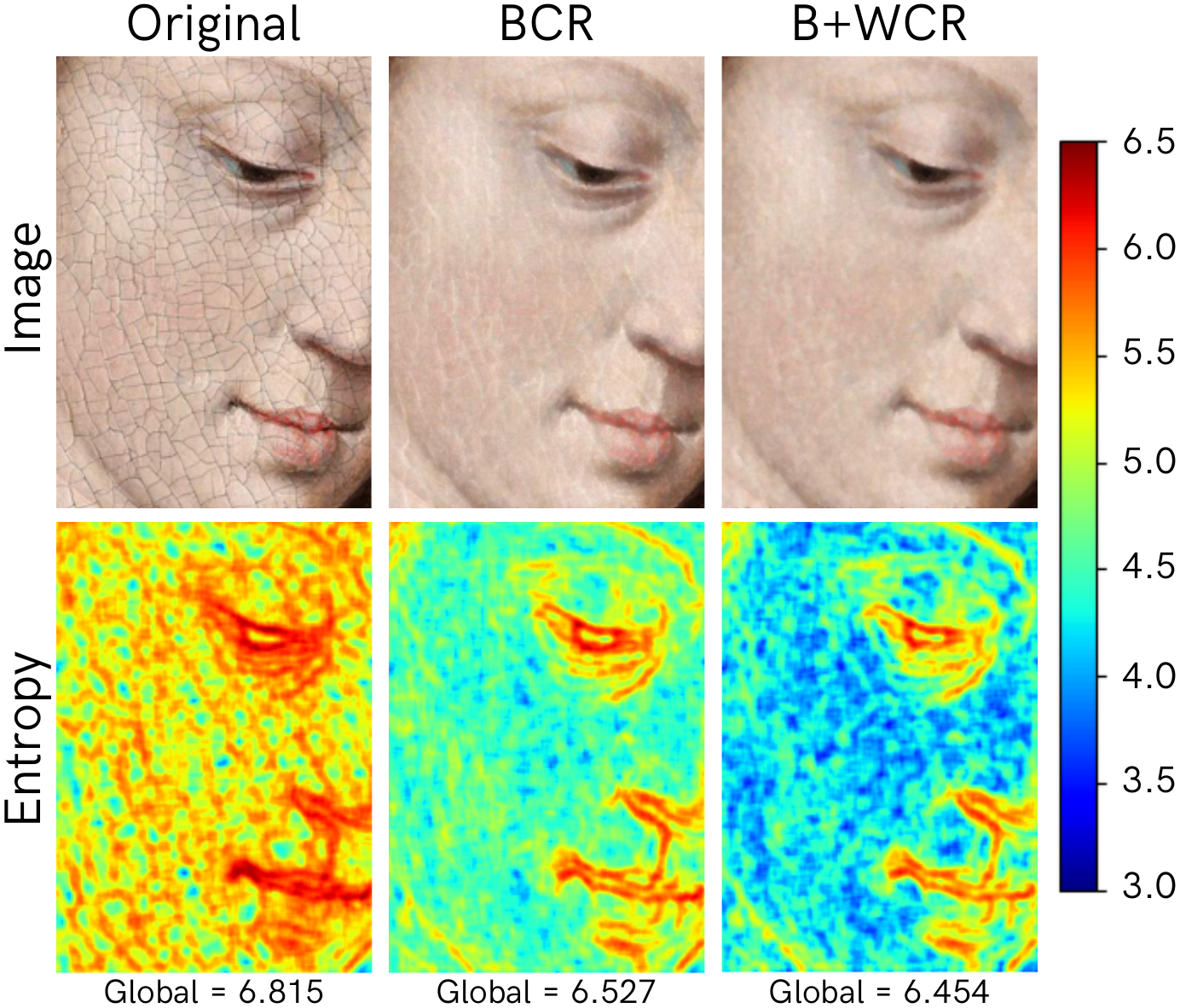}
\vspace{-0.2cm}
\caption{\textbf{Global entropy comparison using the top-hat transform}, evaluating Black Crack Removal (BCR) versus combined Black and White Crack Removal (B+WCR).}
\vspace{-0.2cm}
\label{fig:Entropies}
\end{figure}

\begin{figure}[t]
\centering
% ---- Left: Table ----
\begin{minipage}{\linewidth}
\centering
\captionof{table}{\textbf{Execution time comparison} of crack filling methods with different structuring elements. Times reported in seconds.}
\label{tab:execution_times} \vspace{-0.15cm}

\resizebox{0.7\linewidth}{!}{
\begin{tabular}{lcc}
\toprule
\textbf{Method / SE} & \textbf{Square Kernel} & \textbf{Disk Kernel} \\
\midrule
MTM & 3.2565 & 3.2378  \\
AD & \textbf{3.2015} & \textbf{2.4548} \\
\bottomrule
\end{tabular}
}
\end{minipage}
\vspace{0.2cm}

% ---- Right: Figure ----
\begin{minipage}{\linewidth}
\centering
\includegraphics[width=\linewidth]{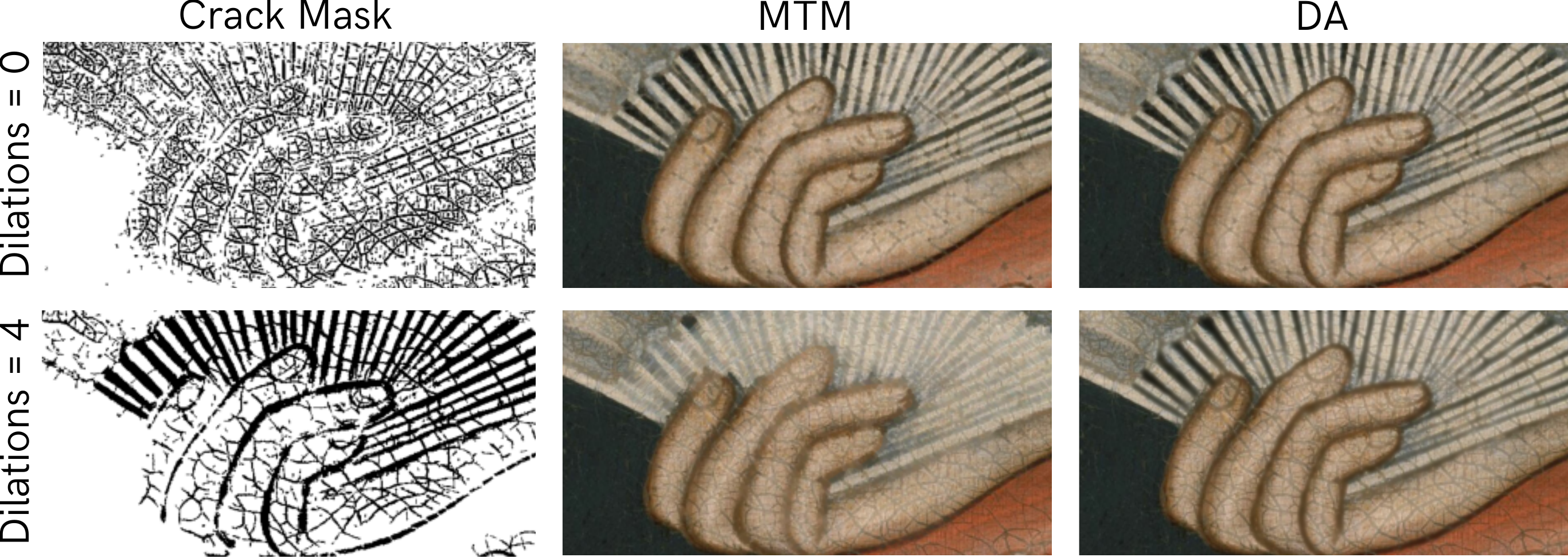}
\vspace{-0.55cm}
\caption{\textbf{Crack-filling comparison } (MTM and AD) under different top-hat dilations; see \cref{sec:crack-detection}.}
\label{fig:MTM_AD}
\vspace{-0.7cm}
\end{minipage}

\end{figure}

The top-hat transform's capability for noise suppression was validated by analyzing global image statistics. This is evidenced by a substantial reduction in global image entropy upon application, particularly when combining both the black and white results (see \cref{fig:Entropies}). This outcome confirms that this transform successfully suppress unnecessary high-frequency components, effectively isolating salient crack features. 

The transform's performance depends on the Structuring Element (SE): we utilized a \(3\times3\) square and a disk of radius 2 pixels, following~\cite{giakoumis2005}. The disk SE was selected for its superior ability to capture diagonal and moderately thicker cracks. Parameter optimization focused on dilation iterations. Minimal dilation (one iteration) was crucial, as additional iterations caused aggressive crack mask expansion and erroneously labeled pictorial elements as cracks. This choice ensures high accuracy in preserving the topology of thin crack lines, as illustrated in \cref{fig:MTM_AD}.

Overall, \cref{fig:Entropies,fig:MTM_AD} qualitative examples showcase the general effectiveness of the proposed pipeline, effectively removing complex craquelure patterns while preserving the integrity of the underlying paint texture.

\subsection{Crack-Filling Methods}
Both MTM and AD are applied solely to pixels labeled as cracks and operate independently per RGB channel before recombination and clipping. While MTM is computationally cheaper, easier to implement, and effective for thin, line-like cracks, AD handles wider cracks and textured areas more robustly. \Cref{tab:execution_times} shows that AD is actually faster for the tested SE, with the disk kernel providing the lowest execution time. Experiments were conducted on an Intel i5-1155G7 CPU  ($2.50 \text{ GHz}$, $16 \text{ GB}$ RAM). Fig.~\ref{fig:MTM_AD} illustrates that using masks with progressively increased dilation blurs fine details in MTM, while AD maintains a sharper structure.

\subsection{Traditional Baseline Comparison}

We benchmarked our automatic pipeline against the traditional, annotation-free Grassfire algorithm~\cite{giakoumis2005}. \Cref{tab:grassfire} presents combined segmentation and inpainting results on our four test images.  Evaluation is based on crack detection metrics accuracy (Acc.) and F$_{1}$ score, and the restoration metric Structural Similarity Index Measure (SSIM). 

The results confirm the efficacy of the learning-based approach. Crucially, the Grassfire method is interactive, requiring time-consuming manual user input for crack separation, whereas our solution provides a fully automatic pipeline with superior performance in all metrics. This validates the automatic suppression of false positives by the SegFormer~\cite{segformer} architecture using synthetic data.

Furthermore, AD consistently outperformed MTM filter for inpainting, yielding the highest restoration quality. AD's ability to diffuse intensity along image isophotes makes it robust for filling wider cracks while preserving structural details, thus confirming it as the optimal filling mechanism.

\begin{table*}[t!]   
\caption{\textbf{Crack detection and restoration performance} using proposed crack detection and filling methods compared with traditional baseline~\cite{giakoumis2005}.}  
\vspace{-0.15cm}
\label{tab:grassfire}
\centering
\resizebox{1\linewidth}{!}{%
\begin{tabular}{c|cc:c|cc:c|cc:c|cc:c}
\toprule
& \multicolumn{6}{c|}{\textbf{Grassfire Algorithm~\cite{giakoumis2005}}} 
& \multicolumn{6}{c}{\textbf{Learning-based Segmentation} (\cref{sec:method})} \\
& \multicolumn{3}{c|}{MTM} & \multicolumn{3}{c|}{AD} 
& \multicolumn{3}{c|}{MTM} & \multicolumn{3}{c}{AD} \\
\cmidrule{2-13}
\multirow{2}{*}{\textbf{Image Name}}  & \multicolumn{2}{c:}{Detection} & Restoration & \multicolumn{2}{c:}{Detection} & Restoration & \multicolumn{2}{c:}{Detection} & Restoration & \multicolumn{2}{c:}{Detection} & Restoration \\
& Acc. $\uparrow$ & F$_{1}$ $\uparrow$ & SSIM $\uparrow$ & Acc. $\uparrow$ & F$_{1}$ $\uparrow$ & SSIM  $\uparrow$ & Acc. $\uparrow$ & F$_{1}$ $\uparrow$ & SSIM $\uparrow$ & Acc. $\uparrow$ & F$_{1}$ $\uparrow$ & SSIM $\uparrow$ \\
\midrule
Image 1 & 65.43 & 26.77 & 63.98 & 67.42 & 29.41 & 67.50 & 75.31 & 36.86 & 71.26 & 78.63 & 39.41 & 77.31 \\ 
Image 2 & 79.76 & 71.50 & 53.45 & 81.74 & 73.54 & 54.05 & 89.74 & 81.82 & 62.18 & 91.91 & 83.54 & 63.07 \\ 
Image 3 & 71.33 & 50.67 & 40.64 & 74.41 & 51.06 & 44.85 & 81.86 & 60.03 & 50.41 & 84.48 & 61.06 & 54.89 \\ 
Image 4 & 70.49 & 49.43 & 54.63 & 72.87 & 51.42 & 57.72 & 82.28 & 59.91 & 61.13 & 85.75 & 61.42 & 64.20 \\
\hline
\textbf{MEAN} & 71.75 & 49.59 & 53.18 & 74.11 & 51.36 & 56.03 & 82.30 & 59.65 & 61.24 & \textbf{85.19} & \textbf{61.36} & \textbf{64.87} \\
\bottomrule
\end{tabular}%
}
\vspace{-0.25cm}
\end{table*}

\begin{table*}[t!]
\centering
\caption{\textbf{Quantitative results}: ablation study of model components and state-of-the-art (SOTA) comparison for crack detection and restoration. \newline \textbf{MCC}: Matthews Correlation Coefficient; \textbf{LPIPS}: Learned Perceptual Image Patch Similarity; \textbf{VIF}: Visual Information Fidelity.}
\vspace{-0.15cm}
\label{tab:quantitative_combined_v2}
\resizebox{\linewidth}{!}{
\setlength{\tabcolsep}{4pt} % Reduced column spacing for fitting
% Adjusted the tabular environment to include a narrow first column for rotated text (15 columns total)
\begin{tabular}{@{}l@{}|lccc|ccccc|ccccc}
\toprule
\multicolumn{1}{c|}{} & \multirow{2}{*}{\hspace{.5cm}\textbf{Method}} & \multicolumn{3}{c|}{\textbf{Ablated Components}} & \multicolumn{5}{c|}{\textbf{Crack Detection Metrics}} & \multicolumn{5}{c}{\textbf{Crack Restoration Metrics}} \\
\cmidrule(lr){3-5} \cmidrule(lr){6-10} \cmidrule(lr){11-15}
\multicolumn{1}{c|}{} & & \textbf{Guided Logit} & \textbf{LoRA} & \textbf{Mask Loss} & Acc.~$\uparrow$ & F1~$\uparrow$ & IoU~$\uparrow$ & Dice~$\uparrow$ & MCC~$\uparrow$ & SSIM~$\uparrow$ & PSNR~$\uparrow$ & MAE~$\downarrow$ & LPIPS~$\downarrow$ & VIF~$\uparrow$ \\
\midrule
% --- Ablation Study Section ---
\multirow{4}{*}{\rotatebox[origin=c]{90}{\textbf{Ablation}}} & SegFormer & -- & -- & -- & 87.44 & 27.64 & 16.04 & 28.54 & 25.82 & 43.40 & 14.18 & 23.33 & 62.30 & 9.36 \\
& \cref{eq:guided-logit} & $\checkmark$ & -- & -- & 85.10 & 28.30 & 26.50 & 35.30 & 33.60 & 49.40 & 15.19 & 28.81 & 59.10 & 4.03 \\
& \cref{eq:lora} & $\checkmark$ & $\checkmark$ & -- & 82.79 & 49.45 & 32.78 & 49.37 & 41.69 & 48.67 & 16.46 & 18.28 & 57.84 & 11.58 \\
& \cref{eq:mask_loss} & $\checkmark$ & $\checkmark$ & $\checkmark$  & \textbf{85.19} & \textbf{61.36} & \textbf{43.86} & \textbf{60.97} & \textbf{58.75} & \textbf{64.87} & \textbf{21.57} & \textbf{12.56} & \textbf{47.56} & \textbf{18.53} \\
\midrule
% --- SOTA Comparison Section ---
\multirow{2}{*}{\rotatebox[origin=c]{90}{\textbf{SOTA\hspace{-0.1cm}}}} & Wan {\em et al.}~\cite{wan2020} & N/A & N/A & N/A & N/A & N/A & N/A & N/A & N/A & 51.35 & 18.06 & 20.61 & 56.87 & 11.83 \\  %82.86 & 49.72 & 34.68 & 51.43 & 45.66
& Pik-Fix~\cite{xu2023} & N/A & N/A & N/A & N/A & N/A & N/A & N/A & N/A & 55.71 & 19.98 & 18.43 & 52.33 & 13.44 \\ %84.37 & 52.63 & 37.57 & 54.72 & 51.75
\bottomrule
\end{tabular}}
\vspace{-0.5cm}
\end{table*}

\subsection{Ablation Study of the Segmentation Modules}

To validate each module's contribution, we present in Table~\ref{tab:quantitative_combined_v2} an ablation study using the AD crack filling algorithm. The results confirm the synergistic effectiveness of our modular approach in boosting crack detection. 

Integrating the detector map into the logits (\cref{eq:guided-logit}) adds a soft prior that yields a modest but consistent improvement. LoRA-based tuning (\cref{eq:lora}) drives the largest performance gain by enabling stable training on limited synthetic data. Finally, the masked loss combined with Dice (\cref{eq:mask_loss}) is critical for suppressing false positives, reaching the highest $\text{F}_{1}=61.36$ and best restoration quality $\text{SSIM}=64.87$.

%The complete model, incorporating all modules, demonstrates how each component incrementally refines the crack mask, contributing to superior final detection and inpainting.

\subsection{Quantitative Comparison}

Given the scarcity of annotation-free models tailored specifically for artwork restoration, we benchmark our approach against the closest adjacent domain: photographic restoration. As detailed in Table~\ref{tab:quantitative_combined_v2}, our method significantly outperforms existing end-to-end restoration models, including Wan {\em et al.} framework~\cite{wan2020} and the specialized inpainting model Pik-Fix~\cite{xu2023}. Zero-shot models like SAM2~\cite{sam2} were tested but failed at effective crack segmentation. Our full pipeline achieves superior performance, securing an SSIM of $64.87$ compared to $55.61$ of Pik-Fix. These results confirm that our detector-guided architecture, combined with efficient synthetic fine-tuning, focuses uniquely on isolating and restoring the fine, irregular characteristics of craquelure.

\vspace{-0.1cm}
\section{Conclusion}
\vspace{-0.1cm}

In this work, we presented an annotation-free framework for the detection and virtual restoration of craquelure in digitized paintings. Addressing the dual challenges of data scarcity and extremely thin crack patterns, our approach bridges the gap between classical and modern image processing. By injecting morphological detection maps into a SegFormer backbone, we successfully created a detector-guided architecture that retains high-frequency structural details while leveraging semantics to suppress false positives.

Our experimental results validate the efficacy of this hybrid strategy, significantly outperforming both traditional interactive methods and state-of-the-art photographic restoration models in a zero-shot setting. The ablation study further highlighted the critical role of LoRA adaptation and masked consistency losses in stabilizing training on synthetic data. Regarding the restoration phase, we demonstrated that AD is superior to MTM filtering.  Collectively, these contributions offer a robust, non-invasive tool for digital conservation, enabling the faithful recovery of fragile artworks without the need for manual pixel-level supervision.

%%%%%%%%%%%%%%%%%%%%%%%%%%%%%%%%%%%%%%%%%%%%%%%%%%%%%%%%%%%%%%%%%%%%%%%%%%%%%%%%

\addtolength{\textheight}{-12cm}   % This command serves to balance the column lengths

%%%%%%%%%%%%%%%%%%%%%%%%%%%%%%%%%%%%%%%%%%%%%%%%%%%%%%%%%%%%%%%%%%%%%%%%%%%%%%%%

%%%%%%%%%%%%%%%%%%%%%%%%%%%%%%%%%%%%%%%%%%%%%%%%%%%%%%%%%%%%%%%%%%%%%%%%%%%%%%%%



\section*{ACKNOWLEDGMENT}

This work has been supported by the project GRAVATAR (PID2023-151184OB-I00), funded by MCIU/AEI/10.13039/501100011033 and the European Regional Development Fund (ERDF), EU.

% The preferred spelling of the word ÒacknowledgmentÓ in America is without an ÒeÓ after the ÒgÓ. Avoid the stilted expression, ÒOne of us (R. B. G.) thanks . . .Ó  Instead, try ÒR. B. G. thanksÓ. Put sponsor acknowledgments in the unnumbered footnote on the first page.

% %%%%%%%%%%%%%%%%%%%%%%%%%%%%%%%%%%%%%%%%%%%%%%%%%%%%%%%%%%%%%%%%%%%%%%%%%%%%%%%%

% References are important to the reader; therefore, each citation must be complete and correct. If at all possible, references should be commonly available publications.

\bibliographystyle{IEEEtran}
\bibliography{references}

\end{document}